\documentclass{article}

\usepackage[preprint]{neurips_2026}

\usepackage[utf8]{inputenc} 
\usepackage[T1]{fontenc}    
\usepackage{hyperref}       
\usepackage{url}            
\usepackage{booktabs}       
\usepackage{amsfonts}       
\usepackage{nicefrac}       
\usepackage{microtype}      
\usepackage{xcolor}         

\usepackage{amsmath, amsthm}
\usepackage{mathtools}
\usepackage{graphicx}
\usepackage{wrapfig}
\usepackage{verbatim}
\usepackage{multirow}

\DeclareMathOperator{\sgn}{sgn}

\title{Non-Expanding Gated Continued Fraction Architecture for Feedforward Layers in Language Models}

\author{%
  Amit Dhurandhar\\
  IBM Research \\
  \texttt{adhuran@us.ibm.com} \\
  \And
  Vijil Chenthamarakshan \\
  IBM Research \\
  \texttt{ecvijil@us.ibm.com} \\
  \And
  Karthikeyan Natesan Ramamurthy \\
  IBM Research \\
  \texttt{knatesa@us.ibm.com} \\
  \AND
  Dennis Wei \\
  IBM Research \\
  \texttt{dwei@us.ibm.com} \\
  \And
  Rahul Nair \\
  IBM Research \\
  \texttt{rahul.nair@ie.ibm.com} \\
}

\begin{document}

\maketitle

\begin{abstract}
Recent work has shown that continued fraction inspired neural architectures (CoFrGeNets) can be highly parameter efficient yet performant for language modeling. In this work, we revisit continued fraction-based channel mixing with the goal of improving its performance and simplifying integration into standard pre-training protocols for language models.
We achieve this by two simple yet critical contributions:
i) we propose a novel non-expanding gated continued fraction inspired architectural component to replace feedforward layers in established language model architectures that may use vanilla multilayer perceptrons or gated linear units or (vanilla/gated) experts in mixture-of-experts (MoE) type architectures. ii) We propose a way to re-implement divisions -- the non-linearity in these architectures -- in terms of logarithms and exponentials. By experimenting on different language modeling architectures such as GPT-2, Llama-3, Mamba-2 and (nano) Llama-MoE we find that our proposed architecture performs similarly or significantly better than the continued fraction architecture proposed in prior work along with having higher throughput. In addition, our re-implementation of divisions makes training of deeper CoFrGeNets stable without the need for incremental training which was proposed in prior work as a means to stabilize training, but added notable complexity to existing large language model training implementations. Code is available at \url{https://github.com/IBM/CoFrNets}.
\end{abstract}

\section{Introduction}

Large Language Models (LLMs) have taken the world by storm \cite{radford2019language, gu2024mamba} revolutionizing how tasks such as web search, writing, coding, planning and even research are being done. A few neural architectures seem to be lying at the core of these advancements. Transformers \cite{attention2017} are arguably the most leveraged architecture appearing in many popular models such as in Generative Pre-trained Transformers (GPTs) \cite{radford2018improving}, Llama \cite{grattafiori2024llama3} and even in the mixture-of-experts type of architectures \cite{dai2024deepseekmoe} which are in vogue currently. State space architectures such as Mamba \cite{gu2024mamba} have gained popularity given that they are more efficient than Transformers especially over long sequences. Diffusion models \cite{diffusion} which leverage Transformers are also gaining traction given their ability to generate sequences in parallel as opposed to auto-regressively. A major and the most parameter heavy component of these architectures is the feedforward network (FFN) which consists of a 2-layer perceptron with the hidden state being expanded by a factor of two or more over the input or embedding dimension. Large number of parameters implies slow and expensive training and inference with a large carbon footprint. Hence, having alternatives that are parameter efficient yet performant is a highly sought after goal \cite{effarchs}.

Recently, a new architecture inspired by the mathematical concept of continued fractions CoFrGeNets \cite{cofrgenets} was proposed for language generation. The main contribution was proposing readily pluggable components for attention and FFNs in transformers that are parameter efficient. Replacing FFNs in dense Transformer models such as GPT2 and Llama led to models that were roughly $\frac{2}{3}$ the size, while replacing both attention and FFNs led to roughly $\frac{1}{2}$ the size with some degradation in performance. A continued fraction can be written as $a_0+\frac{1}{a_1+\frac{1}{a_2+\cdots}}$ (ladder like structure) where, $a_k$s are complex numbers. In the recent CoFrGeNets and the original CoFrNets \citep{cofrnets}, which was introduced for supervised settings, the $a_k$s were replaced by linear functions of the input $x\in \mathbb{R}^p$ by taking the inner product of $x$ with weight vector $w_k\in \mathbb{R}^p$ in each layer $k$.\footnote{Constant term is subsumed in $x$.} The reciprocal of the function thus far is applied as a nonlinearity in each layer leading to the following kind of form for a single ladder: 
\begin{align}
w_0^T x+\frac{1}{w_1^T x+\frac{1}{w_2^T x+\cdots}} \label{eqn:cofrnet}
\end{align}
The $w_k$s are parameters that can be learned. Here, the input $x$ gets multiplied by the parameter vector in each layer with the non-linearity being a reciprocal. This simplistic architecture has nice properties including universal approximability when enough ladders are ensembled together \cite{cofrnets}.

In this work, inspired by CoFrGeNets and motivated by the need to have parameter efficient yet performant alternatives to FFNs which contributes the  most to model size, we propose a novel non-expanding, gated continued fraction inspired architectural component to replace FFNs. The resulting module can be used as a drop-in replacement for FFNs in standard MLP, GLU, and MoE settings, enabling direct integration into architectures such as GPT-2, Llama-style models, and state-space models. This new variant can have slightly more number of parameters than the component proposed in \cite{cofrgenets}, due to the inclusion of the non-expanding gating, but having similar or significantly better performance across models. Additionally, our variant also has higher throughput, a consequence of improved arithmetic intensity.
\begin{figure}[t]
\centering
\includegraphics[width=\textwidth]{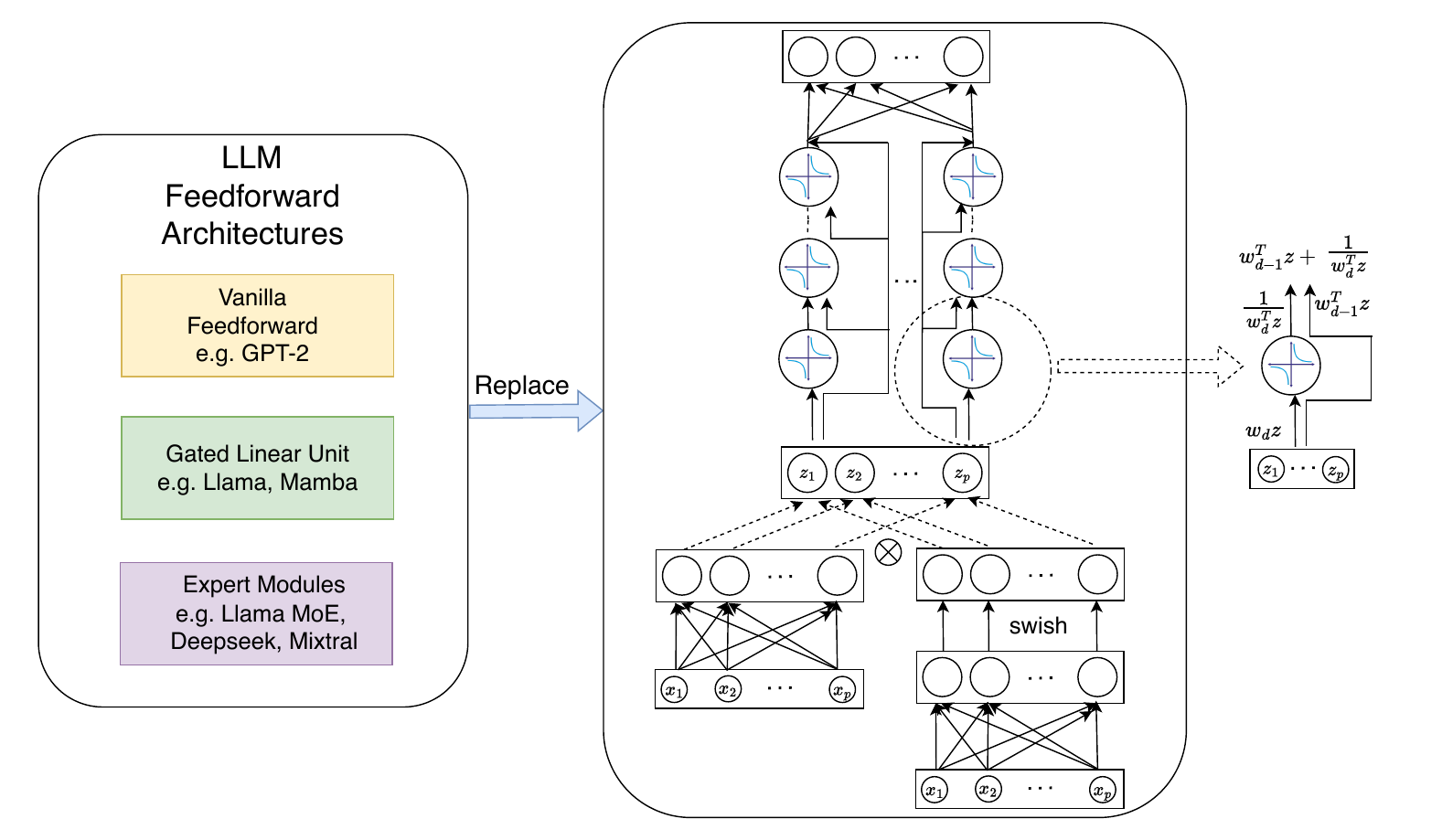}
\caption{Above we see our continued fraction inspired architecture that can replace feedforward layers in dense as well as MoE architectures. Our architecture performs a \emph{non-expansive} gating of the input that then passes through an (ensemble of) continued fraction ladders. The fact that we do not expand the hidden representation and each layer of our ladders requires only a linear number of parameters we obtain significant parameter savings over typical (gated and vanilla) feedforward layers leading to dense models that are $\approx \frac{2}{3}$ the size and MoEs that are $\approx \frac{1}{2}$ the size. In our implementations we create a different class for CoFrGeNet that has the same api as the FFN class, that is, same parameters are passed and the same size output tensor (equal to embedding size) is returned. The depth and width parameters for CoFrGeNet could be additionally passed or just set in the CoFrGeNet class itself. Hence, no changes are made to the residual stream or other aspects of the original architecture. The input $x_1,..., x_p$
 in the figure is thus simply the output of the attention or SSM layer where the residual stream is added and the representation normalized in the particular block. In other words, nothing else in the model architectures changes and the input and output tensors are both size embedding dimension.}
\label{fig:moe_replace}
\end{figure}


One challenge with training continued fraction architectures is that deep ladders can result in slow training and inference as well as training instabilities due to reciprocal calculations. To address the former, in \cite{cofrgenets} a ladder was cast in terms of \emph{continuants} --a mathematically equivalent formulation (see Section \ref{sec:dm})-- which reduced the number of divisions from $d$ to $1$ for a $d$ depth ladder. However, training instabilities persisted for deep ladders (i.e. $d>1$) for which the authors introduced an incremental training protocol whereby training starts at a single depth and incrementally larger depths are trained following a dyadic schedule. This however, added non-trivial complexity to standard pre-training recipes for LLMs. To allow for a drop-in replacement that would require no change to standard pre-training recipes, in this work, we propose a simple reformulation that entirely avoids these stability issues. The reformulation involves transforming the division-based nonlinearities using logarithmic and exponential transformations. By avoiding explicit division, we improve gradient flow leading to improved stability.

As such, our contributions in this work are threefold:

\begin{itemize}
    \item We propose a novel non-expanding gated continued fraction architecture to replace (gated and vanilla) feedforward layers in LLMs (see Figure \ref{fig:moe_replace}), which offers performance and throughput improvements over previously proposed continued fraction architectures \cite{cofrgenets}.
    \item We propose a simple reformulation of the division operation, which accounts for the non-linearity in CoFrNets, in terms for logarithms and exponentials to make training stable for deeper ladders. This facilitates drop-in replacement with a broad class of language models without the need to modify pre-training strategies. 
    
    \item We replace experts in MoE architectures as well as gated FFNs in Mamba-2 with our novel continued fraction architectural component further distinguishing it from prior work. We showcase the broad efficacy of our architectural component by reporting promising results on diverse LLM architectures such as GPT-2, Llama-3, Mamba-2 and Llama-MoE. Notably, the parameter savings for dense models are close to $33$\% which is already considerable, but for MoEs it reaches even higher to about $50$\%.
\end{itemize}

Overall, our contributions retain the lower parameter count and compactness of continued-fraction architectures, while simplifying (stable) training and improving performance in many cases with also improved throughput.




\section{Related Work}
Modern large language models are predominantly built on the transformer backbone~\cite{attention2017}, in which each block alternates a self-attention sublayer with a position-wise feedforward network (FFN). The FFN is typically implemented as a multilayer perceptron with a pointwise nonlinearity, typically ReLU~\cite{nair2010rectified}, GELU~\cite{hendrycks2016gelu} or Swish/SiLU~\cite{ramachandran2017searching}. Recently, the gated linear unit family~\cite{dauphin2017language} was shown to outperform vanilla MLPs at matched parameter budgets~\cite{shazeer2020glu} in several settings, motivating the adoption of SwiGLU in the Llama family of models~\cite{touvron2023llama,touvron2023llama2,grattafiori2024llama3}. A parallel line of work is mixture-of-expert (MoE) models, where each token is routed to a small set of experts that themselves are MLPs or gated MLPs~\cite{shazeer2017outrageously,lepikhin2021gshard,fedus2022switch,jiang2024mixtral,dai2024deepseekmoe,zhu2024llamamoe}. Beyond attention-based models, state-space architectures such as Mamba~\cite{gu2024mamba} and Mamba-2 \cite{dao2024transformers} replace quadratic attention with selective recurrences but still rely on MLP- or GLU-style channel mixers. Across all of these designs, the FFN sublayer dominates the parameter count,  making it a natural target for architectural redesign.

Because of this reason, few works have explored alternatives to dense MLPs. Kolmogorov--Arnold Networks (KANs)~\cite{liu2024kan} replace fixed activations with learnable univariate functions and have shown promise on small-scale scientific tasks in terms of performance and interpretability, although their scalability and training cost remain open questions. Many times KANs are $10$x slower to train than MLPs of the same size. Structured-matrix approaches such as Monarch~\cite{dao2022monarch} factorize dense weights into block-diagonal components to reduce FLOPs and memory, and are not really a novel architecture but a strategy than can be applied across architectures. Polynomial networks~\cite{chrysos2020pinets} introduce multiplicative interactions between linear projections and have been shown effective in vision tasks. 
These alternatives however, have not been shown to match the empirical performance, scalability and stability of MLPs across varied large language model architectures.

The closest work to ours is the recent work CoFrGeNets \cite{cofrgenets}, which proposed continued fraction architectures for attention and FFNs in dense Transformers. The FFN replacements produced models that were $\frac{2}{3}$ the size and with higher throughput. These replacements were however, not incorporated into other models such as Mamba and MoEs. In this work, we incorporate our component into these architectures as well and show that it should be the preferred alternative as we obtain similar level of parameter savings but much better performance in many cases and even higher throughput.

\section{Background and Methods}
\label{sec:dm}

We first introduce some notation. We denote the input or embedding dimension by $p$, the number of ladders in an ensemble by $L$, the sequence length by $l$ and the depth of a ladder by $d$.

Next, we describe what a continued fraction is. The generalized form for a continued fraction is $a_0+\frac{b_1}{a_1+\frac{b_2}{a_2+\cdots}}$, where $a_k$s and $b_k$s can be complex numbers. If none of the $a_k$ or $b_k$ are zero $\forall k\in \mathbb{N}$, then using equivalence transformations \citep{cfbook}, one can create simpler equivalent forms where either the $b_k=1$ or the $a_k=1$ $\forall k\in \mathbb{N}$, with $a_0=0$ in the latter form. A more concise way to write these two forms is as follows: i) $a_0+\frac{1}{a_1+\frac{1}{a_2+\cdots}}\equiv a_0+ \frac{1}{a_1+}\frac{1}{a_2+\cdots}$ and ii) $\frac{b_1}{1+\frac{b_2}{1+\cdots}} \equiv \frac{b_1}{1+} \frac{b_2}{1+\cdots}$. Form i) is known as the \emph{canonical form}.  One of the nice properties of continued fractions is that 
they are the best possible rational approximation to real numbers \citep{cfbook,cfchap}. 

In this work, we consider continued fractions in canonical form, with partial numerators $b_k = 1$ for $k = 1, \dots, d$ and depth $d$. We thus view continued fractions as functions $f$ of the partial denominators, where we separate $a_0$ from the others and use $a \coloneqq (a_1, \dots, a_d)$ as a shorthand. Hence, 
\begin{align}\label{eqn:CFcanon}
f(a_0, a) = a_0 + \frac{1}{a_1 +} \frac{1}{a_2 + } \cdots \frac{1}{a_{d-1} +} \frac{1}{a_d} = a_0 + \tilde{f}(a), 
\end{align}
where we also define $\tilde{f}(a)$ as the ``fractional part'' of $f(a_0, a)$. Another way of representing a continued fraction is in terms of \emph{continuants}, which we describe next. The continued fraction in \eqref{eqn:CFcanon} can be expressed as the following ratio of polynomials $K_{d+1}$ and $K_{d}$,
\begin{equation}\label{eqn:CFcontRatio}
f(a_0, a) = \frac{K_{d+1}(a_0, \dots, a_d)}{K_d(a_1, \dots, a_d)}.
\end{equation}
Polynomials $K_{d}$, $K_{d+1}$ are part of a sequence of polynomials $K_k$, $k = 0, 1, \dots$, known as \emph{continuants}. They satisfy the recursion
\begin{align}
    K_0 = 1, \qquad K_1(a_d) &= a_d,\label{eqn:initCont}\\
    K_k(a_{d-k+1}, \dots, a_d) &= a_{d-k+1} K_{k-1}(a_{d-k+2}, \dots, a_d) \nonumber\\&+ K_{k-2}(a_{d-k+3}, \dots, a_d).\label{eqn:recurCont}
\end{align}
Using \eqref{eqn:recurCont}, \eqref{eqn:CFcontRatio} can also be written as $f(a_0, a) = a_0 + \frac{K_{d-1}(a_2, \dots, a_d)}{K_d(a_1, \dots, a_d)}$ and hence,
\begin{align}
\tilde{f}(a) = \frac{K_{d-1}(a_2, \dots, a_d)}{K_d(a_1, \dots, a_d)}.\label{eqn:CFcontRatio2}
\end{align}

This formalism as shown in previous work \cite{cofrgenets} significantly enhances training and inference speed over standard layerwise forward and backward passes done in \cite{cofrnets}, since the number of divisions for a depth $d$ ladder reduces from $d$ to $1$ and divisions are significantly more expensive than multiplications or additions on modern hardware. Nonetheless, it still causes some training instability especially for deep ladders (even $d\ge 2$). This resulted in the proposal for incremental training requiring non-trivial modifications to language model pre-training procedures. Below we propose a simple alternative that stabilizes training and absolves the need for these training modifications.


We design neural networks inspired by continued fractions by letting the partial denominators $a_k$ be affine functions of an input $x$, $a_k = w_k x$, where $w_k$ is a row vector and a $1$ is prepended to the elements of $x$ so that the corresponding coefficient $w_{k0}$ is the intercept or \textit{bias} term. We will often refer to a continued fraction with $a_k = w_k x$ as a (CoFrNet) \textit{ladder}, and we will also construct ensembles of such ladders.

\subsection{Continued Fraction Architecture for FFNs}

In Figure \ref{fig:moe_replace} we observe our continued fraction architecture for FFNs. We pass the $p$-variate input to two linear layers where the output of one of them (bottom right portion in the figure) is gated by a Swish non-linearity. Then the outputs of the two layers are element-wise multiplied producing a (potentially) sparse representation $(z_1,..., z_p)$. This representation is then input to $L$ continued fraction ladders that produce an $L$-variate output. Note that one could have an arbitrary number of ladders in the ensemble and one projects to $p$ dimensions using the linear layer. The input to the ladders is a gated non-expanded (i.e. the expansion factor $\alpha=1$) representation. The purpose of expansion and sparsification is typically understood to unpack information in a representation \cite{expsparsify}. CoFrNets being a compact yet expressive function class are able to extract information with just partial unpacking, where they benefit from sparsification but do not necessarily need the expansion. Not performing expansion produces significant parameter savings as seen in the experiments. In particular, the number of parameters is $Lp(d+1)+2p^2$ as opposed to the typical $2\alpha p^2$ for vanilla FFNs and $3\alpha p^2$ for gated FFNs, where $\alpha \ge 2$.

The continued fraction architecture suggested in \cite{cofrgenets} and depicted in Figure \ref{fig:mlp} in the appendix consists of two continued fraction ladder ensembles, which directly ingest the input. The output of the ladders is element-wise multiplied for better representation power. Thus no gating of the input was performed and two rather than one ensemble of continued fraction ladders was leveraged. The number of parameters for this variant was $2Lp(d+1)$.

Although our new architecture has a quadratic dependence on $p$, the non-expansion of the intermediate (hidden) FFN layer leads to significant parameter savings in practice as seen for the different models in the experiments. Additionally, as seen in these experiments the number of parameters is only slightly more or similar to the previous continued fraction architecture. More interestingly, even though our architecture may sometimes have more parameters than the previous one, since we have only one ensemble rather than two the \emph{arithmetic intensity is higher} for our model as the input does not have to be read twice and processed separately by two ensembles. The gating is implemented so that the input is projected by a linear map to a representation twice its size and then Swish is applied to the latter half of this representation followed by elementwise multiplication with the prior half needing the input to be read just once. This leads to higher throughput as seen in the experiments, thus mitigating the adverse effects of (the slightly) increased parameter count.

\subsection{Replacing Divisions}
\label{sec:rd}

The continuant formulation described in the previous section does reduce the number of divisions significantly as a $d$-depth ladder which would normally require $d$ divisions now requires only one. Nonetheless, the number of divisions scale with the size of the ensemble requiring $L$-divisions for $L$ ladders. More importantly though even a single division can cause poles or abrupt discontinuities that can hurt gradient based optimization. This was observed in \cite{cofrgenets} and hence for higher depths incremental training was introduced where one would freeze deeper layers and start training them only after the shallower layers were trained reasonably well for the training to be stable.

We propose here a pytorch favorable replacement for the division operation which is performed in CoFrGeNets. 
In particular, the non-linearity in CoFrGeNets is division which is the ratio of two continuants $\frac{K_{d-1}(w_2x, \dots, w_dx)}{K_d(w_1x, \dots, w_dx)}$ for a $d$ depth ladder.
To avoid poles the denominator was capped from below to a small $\epsilon >0$ producing the following non-linearity:
\begin{align}
\label{eqn:div}
    \sgn\left(K_d(w_1x, \dots, w_dx)\right)\frac{K_{d-1}(w_2x, \dots, w_dx)}{\max(|K_d(w_1x, \dots, w_dx)|, \epsilon)}
\end{align}
where $|.|$ denotes absolute value and $\sgn(.)$ the sign function which returns a value of $+1$ or $-1$. We rewrite (\ref{eqn:div}) as follows:

\begin{align}
\label{eqn:nodiv}
    &\sgn(K_{d-1}(w_2x, \dots, w_dx))\sgn\left(K_d(w_1x, \dots, w_dx)\right)\cdot\nonumber\\
    &\exp(\log(|K_{d-1}(w_2x, \dots, w_dx)|)-\log(\max(|K_d(w_1x, \dots, w_dx)|,\epsilon))
\end{align}

This simple re-implementation absolves the need for incremental training. The logarithms and exponential smoothen out the division operation as logarithms compress the dynamic range of values, which reduces the skew in gradients and helps the optimizer move more smoothly towards a minima. This is confirmed in our experiments where CoFrGeNet-FNG is effectively optimized by optimizers such as AdamW for larger ($>1$) depth $d$ and width $L$ of the continued fraction ensemble.

\section{Experiments}
\label{sec:exp}

\subsection{Setup}

\noindent\textbf{Pre-Training Datasets and Models:} We pre-train GPT2-xl and its variants on OpenWebText (OWT) \cite{owt} and the GneissWeb 35B (GW) \cite{gneissweb} datasets. The remaining architectures, that is, Llama-3, Mamba-2 and Llama-MoE (4 experts, 1 activated) are pre-trained on the 2T data mix used to train Docling \cite{Docling}  models. This data mix contains web (DCLM2, DCLM3Plus \cite{li2024datacomplm}), multilingual (FineWeb-2-edu \cite{fineweb-edu}), code (Starcoder, stack-edu \cite{stack-edu-finemath}), math (Finemath \cite{stack-edu-finemath}, Infiwebmath \cite{infimath}, opc-fineweb-math-corpus \cite{opc}) and synthetic data (Cosmopedia \cite{cosmopedia}), which are heavily used to train models for diverse document understanding. The mixture proportions for the different datasets are as follows: \texttt{dclm3plus} $\rightarrow$ 0.5, \texttt{dclm2} $\rightarrow$ 0.24, \texttt{cosmopedia\_v2} $\rightarrow$ 0.06, \texttt{fineweb\_math\_corpus} $\rightarrow$ 0.01, \texttt{finemath-4plus} $\rightarrow$ 0.015, \texttt{infiwebmath-4plus} $\rightarrow$ 0.01, \texttt{finemath-3plus} $\rightarrow$ 0.015, \texttt{infiwebmath-3plus} $\rightarrow$ 0.005, \texttt{starcoder} $\rightarrow$ 0.01, \texttt{stack-edu} $\rightarrow$ 0.17 and \texttt{fineweb-2-edu} $\rightarrow$ 0.01.

\noindent\textbf{Baselines:} We compare our non-expansive gated architecture CoFrGeNet-FNG with i) the original models with the existing FFNs, ii) CoFrGeNet-F which corresponds to replacing FFNs with the architecture proposed in \cite{cofrgenets} and depicted in Figure \ref{fig:mlp} and iii) the original models where non-expansive gating (NG) is applied for the up projection in the FFNs, but where the down projection is a typical (although non-expanded) linear layer (no CoFrNet) seen in the original models. This baseline checks the value of having our continued fraction architecture in the down projection.

\noindent\textbf{Evaluations:} For GPT2-xl and its variants we report perplexity on Penn Tree Bank (PTB) \cite{Marcus93_PTB}, Wikitext2 \cite{Merity2017_Wikitext}, Wikitext103 \cite{Merity2017_Wikitext}, Lambada \cite{Paperno2016_LAMBADA}, AgNews \cite{AgNews2015} and One Billion Words (LM1B) \cite{Chelba2014_lm1b} datasets. We use a stride of 512 for wikitext2, wikitext103 as recommended in these works. For all the other datasets, we use a stride of 256.
We then fine tune our models on GLUE \cite{glue} (classification) tasks and compare accuracies as done in previous works \cite{diffusion, cofrgenets}. We average results over five runs. The other models and their variants we evaluate on openbookqa \citep{OpenBookQA2018}, piqa \citep{piqa}, arc-easy \citep{arc}, winogrande \citep{winogrande}, hellaswag \citep{hellaswag}, lambada open AI \citep{radford2018improving}, boolq \citep{boolq} and sciq \citep{SciQ} which cover open domain Q\&A, reasoning and text understanding tasks. For all the models we also report throughput, train time and floating point operations (FLOPS). For the Llama model we also report NF4 post training quantization (PTQ) results using bitsandbytes (\url{https://huggingface.co/docs/bitsandbytes/en/index}).

We also compare the effect of our division replacement with the previous implementation that performs divisions with and without incremental training. This experiment we report on GPT2-xl as the results are representative for other models which for all the reported experiments leveraged our division re-implementation from section \ref{sec:rd}. We employ a dyadic training schedule for incremental training as recommended in \cite{cofrgenets}.

\noindent\textbf{Parameter Settings:} We use the recommended settings for pre-training GPT2-xl in \url{https://github.com/karpathy/nanoGPT} where, the learning rate is $6\times 10^{-4}$, weight decay is $0.1$, there is no dropout and the maximum iterations is $600K$. 

For fine tuning the GPT2-xl model the learning rate is $0.25\times 10^{-4}$, batch size is $64$ and there is no dropout. This is the same for all the baselines. For our models the learning rate was $0.125\times 10^{-4}$ with other parameters being the same. These learning rates produced the best results for the respective models.

We pre-train the other models for about 2M iterations. The initial learning rate is $3\times 10^{-4}$ and follows an annealing schedule with no dropout.

\begin{wraptable}{r}{.4\textwidth}
\vspace{-1cm}
\scriptsize
\centering
\caption{Below we see architectural details such as number of parameters, number of layers, embedding dimension ($p$) and expansion factor ($\alpha$) of the different (original) models.}
\vspace{2mm}
\begin{tabular}{|c|c|c|c|c|}
  \hline
    \textbf{Model} & \textbf{Size} & \textbf{Layers} & \textbf{$p$} & \textbf{$\alpha$}\\
 \hline\hline
  GPT2-xl & $1.5$B & $48$ & $1600$ & $4$\\
 Llama &$3.2$B & $24$&$3072$&$2.67$\\
  Mamba-2 &$3.2$B&$9$&$4096$&$3.5$\\
  Llama-MoE&$403$M&$30$ &$576$&$2.67$ \\
\hline
\end{tabular}
 \label{tab:archdet}
 \vspace{-0.75cm}
\end{wraptable}

For the continued fraction architectures we set $\epsilon = 0.1$. We experiment with $d$ equal to $1, 2, 5, 8$ with the widths (i.e. number of ladders in each ensemble) taking the same values as well when replacing FFNs. 

All experiments were performed with a batch size of 16 using the AdamW optimizer ($\beta_1=0.9, \beta_2=0.95$) and a weight decay of 0.1. More details of each architecture are given in Table \ref{tab:archdet}.

\begin{table}[htbp]
    \centering
\scriptsize
\centering
\caption{Downstream task accuracies (best results bolded) for GPT2-xl and its variants on GLUE benchmark after finetuning. The first column is the pre-training dataset. Standard deviations are reported in Table \ref{tab:glue_sd} in the appendix. The CoFrGeNet variants perform similarly (best) in this case.}
\vspace{1mm}
\setlength{\tabcolsep}{3pt}
\begin{tabular}
{|c|c|c|c|c|c|c|c|c|c|}
  \hline
    \textbf{Data} &\textbf{Model} & \textbf{MNLI} & \textbf{QQP}& \textbf{QNLI} & \textbf{SST2} & \textbf{COLA} & \textbf{MRPC} & \textbf{RTE} & \textbf{WNLI}\\
 \hline\hline
 \multirow{4}{*}{OWT} & GPT2-xl \tiny{(1.5B)}& $86.89$&$88.93$ &$91.35$&$93.56$&$81.78$&$79.83$&$60.27$&$58.28$\\
 &CoFrGeNet-F \tiny{(972M)}&$\bf 87.24$&$\bf 89.95$ &$\bf 91.87$&$\bf 94.12$&$\bf 82.57$&$\bf 80.17$&$\bf 61.36$&$\bf 58.30$\\
&CoFrGeNet-FNG \tiny{(985M)}&$\bf 87.26$&$\bf 89.95$ &$\bf 91.89$&$\bf 94.16$&$\bf 82.59$&$\bf 80.21$&$\bf 61.35$&$\bf 58.30$\\
&GPT2-xl-NG \tiny{(985M)}&$81.17$&$81.35$ &$83.94$&$87.22$&$75.99$&$74.37$&$56.42$&$58.28$\\
 \hline
 \multirow{4}{*}{GW} & GPT2-xl \tiny{(1.5B)}& $78.28$&$86.83$&$\bf 82.93$&$91.82$&$74.18$&$77.72$&$60.19$&$\bf 58.33$\\
 &CoFrGeNet-F \tiny{(972M)}&$\bf 79.23$&$\bf 87.24$&$82.69$&$\bf 92.38$&$\bf 74.79$&$\bf 78.04$&$\bf 61.37$&$\bf 58.33$\\
 &CoFrGeNet-FNG \tiny{(985M)}&$\bf 79.62$&$\bf 87.26$&$82.73$&$\bf 92.36$&$\bf 74.83$&$\bf 78.01$&$\bf 61.35$&$\bf 58.33$\\
  &GPT2-xl-NG \tiny{(985M)}&$74.83$&$84.74$&$77.82$&$83.94$&$68.93$&$71.83$&$57.85$&$\bf 58.33$\\
  \hline
  \end{tabular}
  \vspace{-1mm}
 \label{tab:glue}
\end{table}

\begin{table}[htbp]
\scriptsize
\centering
\caption{Perplexities of GPT2-xl and its variants with best results bolded.}
\vspace{1mm}
\begin{tabular}{|c|c|c|c|c|c|c|c|}
  \hline
    \textbf{Data} &\textbf{Model} & \textbf{PTB} & \textbf{Wikitxt2}& \textbf{Lbda} & \textbf{AgNews} & \textbf{Lm1b} & \textbf{Wikitxt103}\\
 \hline\hline
 \multirow{4}{*}{OWT} & GPT2-xl \tiny{(1.5B)}&$30.12$ &$18.30$ &$8.66$&$37.13$& $41.20$ & $17.50$\\
&CoFrGeNet-F \tiny{(972M)}&$29.94$ &$\bf 17.09$& $8.15$&$35.72$& $\bf 40.11$ & $16.17$\\
 &CoFrGeNet-FNG \tiny{(985M)}&$\bf 29.87$ &$17.14$& $\bf 8.11$&$\bf 35.69$& $40.17$ & $\bf 16.13$\\
 &GPT2-xl-NG \tiny{(985M)}&$35.98$ &$27.74$& $16.35$&$50.75$& $51.95$ & $29.29$\\
 \hline
 \multirow{4}{*}{GW} & GPT2-xl \tiny{(1.5B)}&$\bf 29.07$ &$19.12$&$31.78$ &$45.62$&$52.36$&$18.93$\\
 &CoFrGeNet-F \tiny{(972M)}&$29.83$ &$\bf 18.08$&$\bf 30.55$ &$41.77$&$\bf 46.59$&$18.11$\\
 &CoFrGeNet-FNG \tiny{(985M)}&$29.68$&$18.10$&$\bf 30.55$ &$\bf 41.65$&$46.86$&$\bf 18.06$\\
 &GPT2-xl-NG \tiny{(985M)}&$34.26$ &$27.38$&$39.44$ &$61.81$&$58.64$&$29.02$\\
 \hline
  \end{tabular}
 \label{tab:ppl}
 \vspace{-0.1cm}
\end{table} 

\noindent\textbf{Hardware:} The GPT2 models were pre-trained on $16$ H100 GPUs leveraging just distributed data parallel (ddp) training as done previously. Fine tuning was done using a single A100 GPU for each model. The Llama-3 and Mamba-2 models were pre-trained using $128$ H100 GPUs with fully sharded distributed data parallel (fsdp) training. The Llama-MoE models were pre-trained using $64$ H100 GPUs with fsdp.

\begin{table}[htbp]
\scriptsize
\centering
\caption{Zero-shot accuracies on open domain Q\&A, reasoning and text understanding tasks for Llama and its variants. As can be seen CoFrGeNet-FNG is much better than CoFrGeNet-F and Llama-NG in most cases and much closer to the original model Llama being roughly $\frac{2}{3}$ the latters size. The docling data mix of $2$ trillion tokens was used for pre-training.}
\vspace{1mm}
\begin{tabular}{|c|c|c|c|c|c|c|c|c|}
  \hline
    \textbf{Model} & \textbf{openqa} & \textbf{piqa}& \textbf{arc} & \textbf{wino} & \textbf{hswag} & \textbf{lambada} & \textbf{boolq} & \textbf{sciq}\\
 \hline\hline
  Llama \tiny{(3.2B)}& $.282$&$.76$ &$.77$&$\bf .654$&$\bf .503$&$.581$&$\bf .691$&$.941$\\
 CoFrGeNet-F \tiny{(1.9B)}&$.292$&$\bf .764$ &$.765$&$.643$&$.482$&$.581$&$.659$&$.941$\\
 CoFrGeNet-FNG \tiny{(2.1B)}&$\bf .294$&$\bf .764$ &$\bf .778$&$.649$&$.491$&$\bf .583$&$.668$&$\bf .944$\\
 Llama-NG \tiny{(2.1B)}&$.277$&$.755$ &$.762$&$.639$&$.477$&$.576$&$.660$&$.938$\\
\hline
   \end{tabular}
  \vspace{-1mm}
 \label{tab:docling-llama}
\end{table}

\begin{table}[htbp]
\scriptsize
\centering
\caption{Zero-shot accuracies on open domain Q\&A, reasoning and text understanding tasks for Mamba-2 and its variants. Here too CoFrGeNet-FNG is much better than CoFrGeNet-F and Mamba2-NG in most cases and closer to the original model Mamba-2 being roughly $\frac{2}{3}$ the latters size. The docling data mix of $2$ trillion tokens was used for pre-training.}
\vspace{1mm}
\begin{tabular}{|c|c|c|c|c|c|c|c|c|}
  \hline
    \textbf{Model} & \textbf{openqa} & \textbf{piqa}& \textbf{arc} & \textbf{wino} & \textbf{hswag} & \textbf{lambada} & \textbf{boolq} & \textbf{sciq}\\
 \hline\hline
Mamba-2 \tiny{(3.2B)}& $\bf .324$&$.761$ &$.768$&$.615$&$\bf .486$&$\bf .548$&$\bf .655$&$.919$\\
 CoFrGeNet-F \tiny{(1.9B)}&$.302$&$.751$ &$.765$&$.613$&$.482$&$.541$&$.651$&$.913$\\
 CoFrGeNet-FNG \tiny{(2.1B)}&$.322$&$\bf .763$ &$\bf .773$&$\bf .619$&$.484$&$.543$&$.651$&$\bf .924$\\
 Mamba-2-NG \tiny{(2.1B)}&$.297$&$.757$ &$.762$&$.609$&$.478$&$.539$&$.646$&$.918$\\
\hline
   \end{tabular}
  \vspace{-1mm}
 \label{tab:docling-mamba}
\end{table}

\begin{table}[htbp]
\scriptsize
\centering
\caption{Zero-shot accuracies on open domain Q\&A, reasoning and text understanding tasks for Llama-MoE and its variants. In this case, CoFrGeNet-FNG is much better than CoFrGeNet-F, Llama-MoE-NG and even the original model Llama-MoE in most cases being roughly half the latters size. The docling data mix of $2$ trillion tokens was used for pre-training.}
\vspace{1mm}
\begin{tabular}{|c|c|c|c|c|c|c|c|c|}
  \hline
    \textbf{Model} & \textbf{openqa} & \textbf{piqa}& \textbf{arc} & \textbf{wino} & \textbf{hswag} & \textbf{lambada} & \textbf{boolq} & \textbf{sciq}\\
 \hline\hline
Llama-MoE \tiny{(403M)}& $.142$&$.6$ &$.472$&$\bf .501$&$.275$&$.178$&$\bf .608$&$.726$\\
 CoFrGeNet-F \tiny{(204M)}&$.148$ &$.609$&$\bf .511$&$.499$&$.280$&$.196$&$.414$&$.761$\\
 CoFrGeNet-FNG \tiny{(204M)}&$\bf .152$&$\bf .610$ &$\bf .511$&$.498$&$\bf .281$&$\bf .212$&$.532$&$\bf .776$\\
 Llama-MoE-NG \tiny{(204M)}&$.142$& $.576$&$.417$&$.499$&$.267$&$.157$&$.607$&$.681$\\
\hline
   \end{tabular}
  \vspace{-1mm}
 \label{tab:docling-moe}
\end{table}

\subsection{Observations}

\noindent\textbf{Performance:} From Tables \ref{tab:glue} and \ref{tab:ppl} we see that our CoFrGeNet-FNG architecture performs best or close to it on different benchmarks for GPT2-xl. It has slightly more parameters than the previously suggested CoFrNet architecture CoFrGeNet-F. Since, GPT2-xl employs a vanilla FFN it seems the gating in our architecture is not as needed, although it does not harm the performance either. Not using a CoFrNet and doing non-expansive gating however, hurts performance. 

Looking at Tables \ref{tab:docling-llama} and \ref{tab:docling-mamba} we see the advantage of our architecture where it is closer to the original Llama and Mamba-2 models than the CoFrGeNet-F. It is in fact significantly better than CoFrGeNet-F on many benchmarks and where it is not worse than it even in a single instance. The non-expansive gated version without CoFrNet is again worse. This shows our architecture is particularly useful for models that use gated FFNs.
\begin{wrapfigure}{r}{.8\textwidth}
\vspace{-.25cm}
\centering
\includegraphics[width=.8\textwidth]{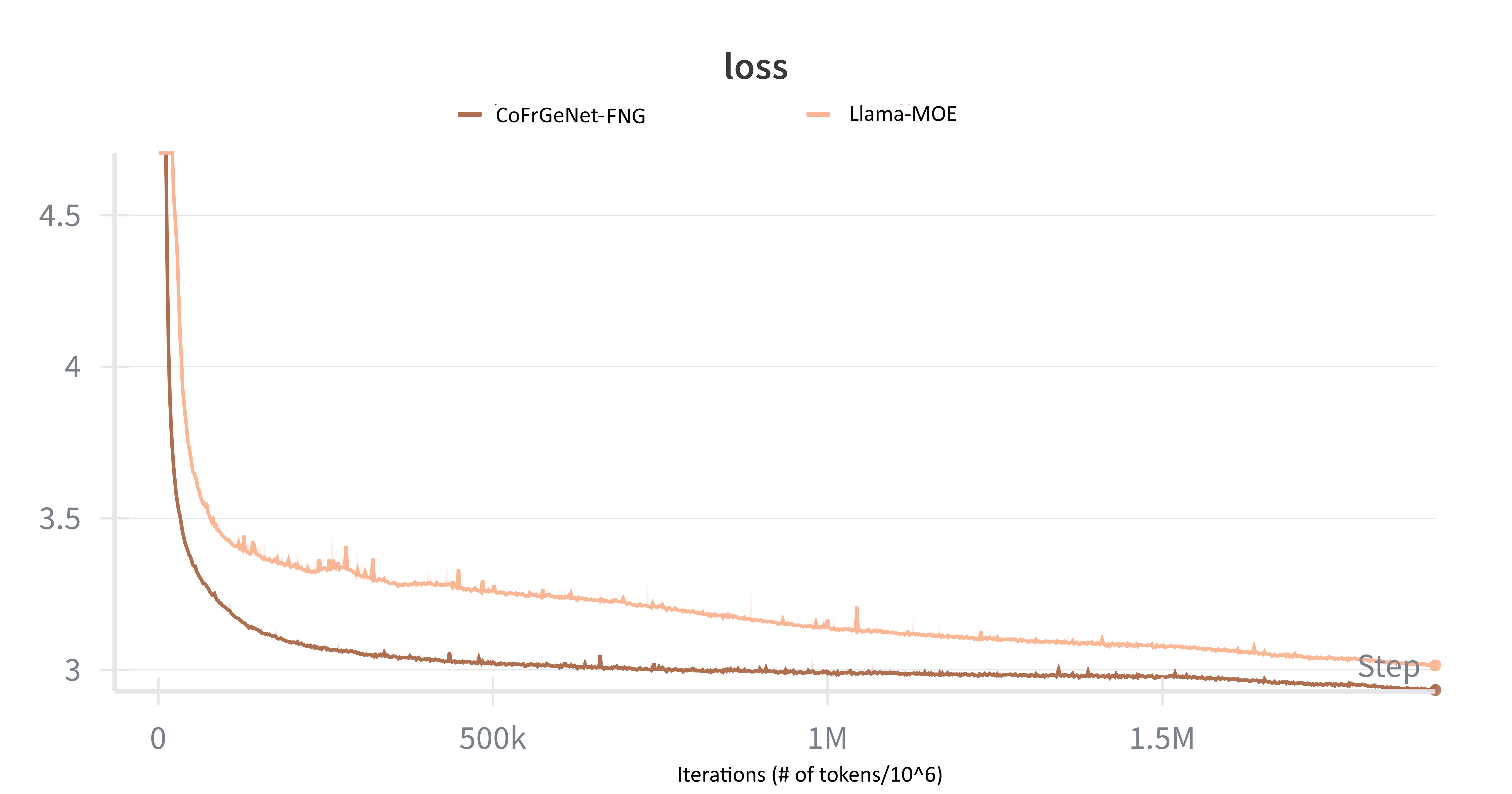}
\caption{Validation loss for Llama-MoE and CoFrGeNet-FNG as a function of the number of training tokens. As can be seen the loss for CoFrGeNet-FNG converges faster and to a lower value ($2.92$ vs $3.01$) than the original model qualitatively consistent with the benchmark results in Table \ref{tab:docling-moe}.}
\label{fig:moe}
\end{wrapfigure}

For the MoE case, we observe in Table \ref{tab:docling-moe} that the non-expansive versions are roughly half the size of the original MoE, and our model outperforms even the original model in most cases. Here the size becomes roughly half because more (4 in this case) FFNs (a.k.a. experts) are replaced by our non-expansive gated architecture. As the number of experts increase, our parameter savings should approach the expansion factor in these models as a larger fraction of the model would be our continued fraction architecture. For Llama-MoE the expansion factor was 2.67 and so we start coming close to it in terms of scale of parameter savings as we move from dense models to MoEs. Hence, with more experts the percentage savings in parameters will be even higher. Interestingly here, the new architecture has the same size as CoFrGeNet-F. Also observing the validation loss during pre-training in Figure \ref{fig:moe}, we see faster convergence and better loss when compared to the original Llama-MoE.

\noindent\textbf{Efficiency:} In Table \ref{tab:doclingspeed} we observe the throughput, train times and FLOPS for the different models. Our variant, CoFrGeNet-FNG, is much faster than the original models and requires much fewer FLOPS. While CoFrGeNet-FNG although is slightly larger than the previously proposed CoFrGeNet-F variant, in most cases it has higher throughput and lower training time. This is because the (gated) input has to pass only through a single ensemble of ladders as opposed to two distinct ones and hence has to be read only once leading to higher arithmetic intensity for the new architecture. This improves training efficiency albeit with slightly more parameters and FLOPS/token.

\noindent\textbf{Post Training Quantization:} We see in Table \ref{tab:docling-llama-ptq} that the accuracy drop after performing 4-bit PTQ on the CoFrGeNet-FNG variant of Llama is minimal for all benchmarks. This implies that our variants are robust to quantization. The peak memory usage dropped from $8.3548$ GB to $2.3564$ GB for the quantized variant.

\begin{wraptable}{r}{.75\textwidth}
\vspace{-.65cm}
\scriptsize
\centering
\caption{Throughput, training time and the number of (training) floating point operations (FLOPS) for the different models and our variants of them. For GPT2-xl the train time is for OWT, GW respectively, while for others its on the docling data mix.}
\vspace{1mm}
\begin{tabular}{|c|c|c|c|}
  \hline
    \textbf{Model} & \textbf{Tokens/day} & \textbf{Train Time (days)} & \textbf{FLOPS/token (GF)}\\
 \hline\hline
  GPT2-xl \tiny{(1.5B)}& $1.15$B&$7.9$, $17.2$&$9.1$\\
 CoFrGeNet-F \tiny{(972M)}&$1.17$B&$7.8$, $16.9$&$5.9$\\
  CoFrGeNet-FNG \tiny{(985M)}&$1.20$B&$7.6$, $16.2$&$6.1$\\
  GPT2-xl-NG \tiny{(985M)}&$1.21$B&$7.6$, $16.1$ &$6.1$ \\
\hline
\hline
 Llama \tiny{(3.2B)}& $235$B&$8.5$&$19.3$\\
 CoFrGeNet-F \tiny{(1.9B)}&$288$B&$7$&$11.7$\\
  CoFrGeNet-FNG \tiny{(2.1B)}&$306$B&$6.5$&$12.5$\\
  Llama-NG \tiny{(2.1B)}&$306$B&$6.5$&$12.5$ \\
\hline
\hline
 Mamba-2 \tiny{(3.2B)}& $475$B&$4.2$&$19.1$\\
 CoFrGeNet-F \tiny{(1.9B)}&$478$B&$4.2$&$11.5$\\
  CoFrGeNet-FNG \tiny{(2.1B)}&$486$B&$4.1$&$12.2$\\
  Mamba2-NG \tiny{(2.1B)}&$491$B&$4$&$12.1$ \\
\hline
\hline
 Llama-MoE \tiny{(403M)}& $299$B&$6.7$&$1.1$\\
 CoFrGeNet-F \tiny{(204M)}&$301$B&$6.6$&$0.70$\\
  CoFrGeNet-FNG \tiny{(204M)}&$309$B&$6.4$&$0.71$\\
  Llama-MoE-NG \tiny{(204M)}&$310$B&$6.3$&$0.70$ \\
\hline
   \end{tabular}
 \label{tab:doclingspeed}
\end{wraptable}
\noindent\textbf{Replacing Divisions:} Note that all models were trained using equation \ref{eqn:nodiv} as a replacement for divisions. We see from the results we discussed before that the performance for CoFrGeNet-F matches that of those reported in \cite{cofrgenets}, which did not use this replacement, but used incremental training, thus showing the efficacy of our no division replacement. Moreover, the results for our variant CoFrGeNet-FNG is competitive or better than the original models in several cases further strengthening our claim. 

\begin{table}[htbp]
\scriptsize
\centering
\caption{Zero-shot accuracy difference between (bf16) CoFrGeNet-FNG for Llama and its post training quantized NF4 variant. As can be seen the difference in accuracies is minimal on open domain Q\&A, reasoning and text understanding tasks. The docling data mix of $2$ trillion tokens was used for pre-training.}
\vspace{1mm}
\begin{tabular}{|c|c|c|c|c|c|c|c|c|}
  \hline
    \textbf{Model} & \textbf{openqa} & \textbf{piqa}& \textbf{arc} & \textbf{wino} & \textbf{hswag} & \textbf{lambada} & \textbf{boolq} & \textbf{sciq}\\
 \hline\hline
 CoFrGeNet-FNG \tiny{(2.1B)}&$.0120$&$.0141$ &$.0126$&$.0032$&$.0120$&$.0136$&$.0141$&$-.0070$\\
\hline
   \end{tabular}
  \vspace{-1mm}
 \label{tab:docling-llama-ptq}
\end{table}

\begin{wraptable}{r}{.75\textwidth}
\vspace{-.5cm}
\scriptsize
\centering
\caption{Perplexities of CoFrGeNet-FNG (GPT2-xl) with no incremental training (NIT), with incremental training (IT) and with no incremental training and using no divisions log/exp trick (NIT-ND). NIT-ND performs similar to IT absolving the need for complicating the standard training procedure of language models. Best results bolded.}
\vspace{1mm}
\begin{tabular}
{|c|c|c|c|c|c|c|c|}
  \hline
    \textbf{Data} &\textbf{Model} & \textbf{PTB} & \textbf{Wikitxt2}& \textbf{Lbda} & \textbf{AgNews} & \textbf{Lm1b} & \textbf{Wikitxt103}\\
 \hline\hline
 \multirow{3}{*}{OWT} 
&NIT &$33.72$ &$26.71$& $12.56$&$42.18$& $47.28$ & $22.65$\\
 &IT &$29.89$&$\bf 17.12$& $8.12$&$35.72$& $\bf 40.14$ & $16.14$\\
 &NIT-ND &$\bf 29.87$ &$17.14$& $\bf 8.11$&$\bf 35.69$& $40.17$ & $\bf 16.13$\\
 \hline
 \multirow{3}{*}{GW} 
 &NIT&$35.88$ &$25.55$&$37.33$ &$45.46$& $49.53$& $20.44$\\
  &IT&$29.72$ &$18.13$&$\bf 30.52$ &$\bf 41.63$&$\bf 46.83$&$18.11$\\
 &NIT-ND&$\bf 29.68$&$\bf 18.10$&$30.55$ &$41.65$&$46.86$&$\bf 18.06$\\
 \hline
  \end{tabular}
 \label{tab:rd}
 \vspace{-0.1cm}
\end{wraptable}
In Table \ref{tab:rd}, we see a comparison of the three approaches --no incremental training (NIT) using divisions, incremental training (IT) using divisions, and no incremental training and no divisions (NIT-ND)-- for depth $5$ CoFrGeNet-FNG networks replacing FFNs in GPT2-xl. As can be seen the perplexities are quite similar for IT and NIT-ND with NIT being much worse. This apples-to-apples comparison clearly shows that our reformulation is effective in stabilizing the training of these networks foregoing the need for incremental training.

\section{Discussion}
\label{sec:disc}

In this work we have proposed a novel continued fraction architecture to replace FFNs that performs similarly or significantly better than prior work, while still being parameter efficient and having higher throughput. We have evaluated this across different models such as GPT2, Llama-3, Mamba-2 and Llama-MoE, where the latter two replacements were not performed in prior work. We also proposed a simple way to re-implement divisions in these architectures in terms of logarithms and exponentials making training stable and accurate precluding the need for incorporating complex incremental training procedures.

Even though our work significantly enhances the efficacy and applicability of continued fraction architectures for language modeling it has certain \emph{limitations}. Our networks are still susceptible to prompt injection attacks, hallucinations and other kinds of output manipulation that other architectures are. A memory efficient implementation possibly leveraging Triton kernels is something missing in our implementations, which would possibly further speed up models. We would also like to test our contributions on larger MoE models, although that is constrained by the amount of compute we have available.

We believe, our work makes an important contribution to the field of efficient architectures which can have \emph{significant societal implications}. Having to train more compact models can lead to savings in time, cost and a reduced carbon footprint. They could be easier to audit, maintain and track. The favorable post training quantization results further support that compact CoFrGeNet models can be performant. Moreover, given that our proposed component has favorable interpretability properties \cite{cofrnets} may open new avenues to appropriate trust in these models rather than relying solely on post-hoc (mechanistic) interpretability methods, which could have varying degrees of faithfulness.

In the future, it would be exciting to understand if there are ways to guide this unique architecture, which has multiple intervention points as it accepts an input at every layer, to inherently become more robust or safe without simply doing the standard data augmented training. Foraying into such and other trust and efficiency related directions would be, in our opinion, a worthwhile endeavor.

\bibliography{ref}
\bibliographystyle{abbrv}

\appendix

\section{Appendix}

\begin{table}[htbp]
    \centering
\scriptsize
\centering
\caption{Downstream task accuracies \emph{with standard deviations} (best results bolded) for GPT2-xl and its variants on GLUE benchmark after finetuning. The first column is the pre-training dataset.}
\vspace{1mm}
\begin{tabular}
{|c|c|c|c|c|c|c|c|c|c|}
  \hline
    \textbf{Data} &\textbf{Model} & \textbf{MNLI} & \textbf{QQP}& \textbf{QNLI} & \textbf{SST2} & \textbf{COLA} & \textbf{MRPC} & \textbf{RTE} & \textbf{WNLI}\\
 \hline\hline
 \multirow{4}{*}{OWT} & GPT2-xl \tiny{(1.5B)}& $86.89$\tiny{$\pm.15$}&$88.93$\tiny{$\pm.67$} &$91.35$\tiny{$\pm.34$}&$93.56$\tiny{$\pm.24$}&$81.78$\tiny{$\pm.38$}&$79.83$\tiny{$\pm.26$}&$60.27$\tiny{$\pm.22$}&$58.28$\tiny{$\pm.28$}\\
 &CoFrGeNet-F \tiny{(972M)}&$\bf 87.24$\tiny{$\pm.14$}&$\bf 89.95$\tiny{$\pm.17$} &$\bf 91.87$\tiny{$\pm.22$}&$\bf 94.12$\tiny{$\pm.11$}&$\bf 82.57$\tiny{$\pm.28$}&$\bf 80.17$\tiny{$\pm.33$}&$\bf 61.36$&$\bf 58.30$\tiny{$\pm.09$}\\
&CoFrGeNet-FNG \tiny{(985M)}&$\bf 87.26$\tiny{$\pm.18$}&$\bf 89.95$\tiny{$\pm.12$} &$\bf 91.89$\tiny{$\pm.34$}&$\bf 94.16$\tiny{$\pm.29$}&$\bf 82.59$\tiny{$\pm.23$}&$\bf 80.21$\tiny{$\pm.19$}&$\bf 61.35$\tiny{$\pm.32$}&$\bf 58.30$\tiny{$\pm.16$}\\
&GPT2-xl-NG \tiny{(985M)}&$81.17$\tiny{$\pm.23$}&$81.35$\tiny{$\pm.35$} &$83.94$\tiny{$\pm.21$}&$87.22$\tiny{$\pm.31$}&$75.99$\tiny{$\pm.42$}&$74.37$\tiny{$\pm.27$}&$56.42$\tiny{$\pm.08$}&$58.30$\tiny{$\pm.19$}\\
 \hline
 \multirow{4}{*}{GW} & GPT2-xl \tiny{(1.5B)}& $78.28$\tiny{$\pm.82$}&$86.83$\tiny{$\pm.17$}&$\bf 82.93$\tiny{$\pm.37$}&$91.82$\tiny{$\pm.22$}&$74.18$\tiny{$\pm.82$}&$77.72$\tiny{$\pm.93$}&$60.19$\tiny{$\pm.01$}&$\bf 58.33$\tiny{$\pm.07$}\\
 &CoFrGeNet-F \tiny{(972M)}&$\bf 79.23$\tiny{$\pm.54$}&$\bf 87.24$\tiny{$\pm.35$}&$82.69$\tiny{$\pm.61$}&$\bf 92.38$\tiny{$\pm.32$}&$\bf 74.79$\tiny{$\pm.25$}&$\bf 78.04$\tiny{$\pm.21$}&$\bf 61.37$\tiny{$\pm.51$}&$\bf 58.33$\tiny{$\pm.44$}\\
 &CoFrGeNet-FNG \tiny{(985M)}&$\bf 79.62$\tiny{$\pm.63$}&$\bf 87.26$\tiny{$\pm.25$}&$82.73$\tiny{$\pm.53$}&$\bf 92.36$\tiny{$\pm.45$}&$\bf 74.83$\tiny{$\pm.56$}&$\bf 78.01$\tiny{$\pm.34$}&$\bf 61.35$\tiny{$\pm.08$}&$\bf 58.33$\tiny{$\pm.04$}\\
  &GPT2-xl-NG \tiny{(985M)}&$74.83$\tiny{$\pm.66$}&$84.74$\tiny{$\pm.37$}&$77.82$\tiny{$\pm.49$}&$83.94$\tiny{$\pm.42$}&$68.93$\tiny{$\pm.29$}&$71.83$\tiny{$\pm.38$}&$57.85$\tiny{$\pm.41$}&$\bf 58.33$\tiny{$\pm.18$}\\
  \hline
  \end{tabular}
  \vspace{-1mm}
 \label{tab:glue_sd}
\end{table}

\begin{figure}[htbp]
\centering
\includegraphics[width=.45\textwidth]{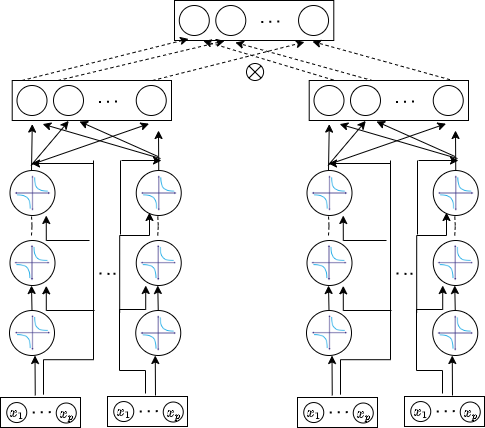}
  \caption{CoFrNet architecture from \cite{cofrgenets} to simulate FFNs. Two ensembles are used each consisting of specified number of $p$-variate ladders. Element wise multiplication is done between the representations coming out of the linear layer of each ensemble for higher expressivity.}
\label{fig:mlp}
\end{figure}



\end{document}